# Dynamic Error-bounded Lossy Compression (EBLC) to Reduce the Bandwidth Requirement for Real-time Vision-based Pedestrian Safety Applications

Mizanur Rahman, Mhafuzul Islam, Jon C. Calhoun, *Member, IEEE,* and Mashrur Chowdhury, *Senior Member, IEEE*

***Abstract*— As camera quality improves and their deployment moves to areas with limited bandwidth, communication bottlenecks can impair real-time constraints of an ITS application, such as video-based real-time pedestrian detection. Video compression reduces the bandwidth requirement to transmit the video but degrades the video quality. As the quality level of the video decreases, it results in the corresponding decreases in the accuracy of the vision-based pedestrian detection model. Furthermore, environmental conditions (e.g., rain and darkness) alter the compression ratio and can make maintaining a high pedestrian detection accuracy more difficult. The objective of this study is to develop a real-time error-bounded lossy compression (EBLC) strategy to dynamically change the video compression level depending on different environmental conditions in order to maintain a high pedestrian detection accuracy. We conduct a case study to show the efficacy of our dynamic EBLC strategy for real-time vision-based pedestrian detection under adverse environmental conditions. Our strategy selects the error tolerances dynamically for lossy compression that can maintain a high detection accuracy across a representative set of environmental conditions. Analyses reveal that our strategy increases pedestrian detection accuracy up to 14% and reduces the communication bandwidth up to 14x for adverse environmental conditions compared to the same conditions but without our dynamic EBLC strategy. Our dynamic EBLC strategy is independent of detection models and environmental conditions allowing other detection models and environmental conditions to be easily incorporated in our strategy.**

## I. Introduction

THE number of pedestrian fatalities has risen each year with over 6,000 reported deaths in 2018 alone, an increase of over 30% compared to 2009 [1]. The presence of a pre-crash warning system, which tracks both vehicles and pedestrian movements, could have prevented most of these pedestrian-related crashes. Addressing the number of traffic fatalities is a matter of national importance [2]. As transportation begins to shift toward autonomous and self-driving vehicles, roadways and intersections are being outfitted with safety devices, such as cameras and sensors to improve pedestrian safety [3, 4]. Even modern vehicles include an in-vehicle vision-based pedestrian warning system to assist drivers in avoiding pedestrian-related [5, 6]. However, in-vehicle pedestrian warning systems do not provide any pre-crash warning to pedestrians.

Vision-based pedestrian detection relies on image classification of frames taken from roadside cameras at signalized intersections [5, 6]. Because there is a limited computational capability near the cameras, the video data must be sent to a roadside transportation data infrastructure (roadside edge computing device) or to the cloud for video processing. As the size of the video increases, so too does the latency to determine if pedestrians are on the roadway. Increasing the latency decreases the likelihood that pedestrians are detected in real-time, and without real-time detection of pedestrians, improvements to safety on roadways will be limited. Moreover, as high-resolution cameras and an increase in the number of connected devices compete for the available bandwidth, the bandwidth available to safety-critical applications may inhibit public safety. Thus, efficient communication of video data for pedestrian detection is required to ensure the safety of pedestrians on or near roadways.

Data compression trades computational time for a reduction in data size. Video compression algorithms employ lossy data compression, which trades inaccuracies in the video's frames for larger reductions in video size [7]. However, as the level of loss increases, the quality of the video decreases. For image and video compression algorithms, this typically results in "blocking", where pixel blocks are approximated by a single value [8]. Quantifying the level of acceptable loss defines the video compression limit for a given algorithm. Common metrics to evaluate the level of loss in video data include peak signal-to-noise ratio (PSNR), root-mean-squared error (RMSE), and structural similarity index (SSIM) [9]. Lossy compression algorithms that ensure a fixed level of loss in the compressed data are referred to as error-bounded lossy compression (EBLC) algorithms.

Through the judicious use of error-bounded lossy compression (EBLC), video streaming companies, such as Netflix and YouTube, optimize the video quality given the amount of available bandwidth [10] In addition, our prior work

M. Rahman, M. Islam and M. Chowdhury are with the Glenn Department of Civil Engineering, Clemson University, Clemson, SC 29634 USA (e-mail: mdr@clemson.edu, mdmhafi@clemson.edu and mac@clemson.edu,).

J. Calhoun is with the Holcombe Department of Electrical and Computer Engineering at Clemson University, Clemson, SC 29634 (e-mail: jonccal@clemson.edu).

has shown the utility of using EBLC for real-time pedestrian detection [5]. In our prior work, researchers were able to reduce the bandwidth requirements to transmit video data by 30x without deterioration in the pedestrian detection accuracy. However, this work considers has several limitations. First, they use a single static error tolerance for their deployment. Next, they feed compressed data into a detection model designed for uncompressed data. Finally, they evaluate their system using a limited number of environmental conditions. However, as we show in Section V, environmental conditions (e.g., rain, darkness) alter the compression ratio and can make it more difficult to maintain a high detection accuracy for pedestrian safety applications. Thus, adapting the error tolerance based upon environmental conditions ensures pedestrian detection accuracy does not deteriorate in adverse environmental conditions.

The objective of this study is to reduce bandwidth requirements for pedestrian detection by developing a real-time EBLC strategy to dynamically change the video compression threshold depending on different environmental conditions while maintaining a high pedestrian detection accuracy. Moreover, we adapt the detection model based on the compression level to improve detection accuracy on highly compressed data. Using this algorithm, we maintain an appropriate pedestrian accuracy across a representative selection of environmental conditions.

The remainder of this paper is broken down into the following sections. Section II describes the contribution of the paper. Section III describes the related work on error bounded lossy compression, pedestrian detection, and environment classification methods. Section IV presents the dynamic error bunded lossy video compression strategy. Section V entails an evaluation of the method and the analytical results. Finally, concluding remarks are provided in Section VI.

## II. Contribution of the Paper

The contribution of our paper is the development of a dynamic error-bounded lossy compression (EBLC) strategy for video feeds used in real-time pedestrian detection. The dynamic compression strategy accounts for environmental factors ensuring a defined pedestrian detection accuracy is maintained while effectively reducing the communication bandwidth requirements. The strategy is independent of the pedestrian detection model such that any pedestrian detection model can be used in the framework presented in this paper. In addition, any other environmental factor, such as snow, can be incorporated in our dynamic EBLC strategy by replicating our steps in Section IV for the new environmental condition. Moreover, our strategy is applicable to image recognition applications beyond pedestrian detection where environmental conditions or the visual quality of video feeds vary overtime. The EBLC strategy can reduce the communication bandwidth usage of a video feed, which allows more videos to be transmitted concurrently through a fixed bandwidth. Furthermore, dynamic EBLC significantly reduces the storage requirements to archive videos for offline analysis. Thus, our EBLC strategy allows storage of more minutes of video without the need of modifying the underlying hardware.

## III. Related Work

This section describes existing work related to error-bounded lossy compression (EBLC), pedestrian detection, and image classification methods. Examining the limitations of the existing methods, we identify an appropriate lossy video compression technique, pedestrian detection, and image classification method for our dynamic EBLC strategy.

### A. Error-bounded lossy compression

Lossless data compression allows for the reduction in data size with no loss in the data's accuracy. Lossy compression (LC) significantly reduces data sizes and offers better compression ratios than lossless compression, but at the expense of inaccuracies in the decompressed data [11]. In the context of video compression, LC compresses by introducing noise into each frame by representing the frame with fewer bits [7]. Typically, the larger the loss in data accuracy, the larger the compression ratio [10, 12]. Current state-of-the-art LC algorithms known as error-bounded lossy compression (EBLC) algorithms offer the ability to control the level of loss introduced when compressing the data [13]. Modern video compression algorithms, such as H.264 [14] and high-efficiency video coding (HEVC) [15] are optimized for high-resolution videos by encoding more information into each compressed bit. H.264 and HEVC compress videos by identifying regions of inter- and intra-frame similarity and then applying transforms, such as the discrete cosine transform [16] and encoding the coefficients or using delta encoding to encode the differences between two frames.

Due to the need to understand the impact of inaccuracies on the quality-of-service, EBLC has not received much attention in the intelligent transportation systems (ITS) domain. In the context of pedestrian detection, quality-of-service is determined by maintaining fixed detection accuracy. Any deterioration in the detection accuracy can lead to unsafe situations for pedestrians. Our prior work [5] shows that using EBLC and a static error tolerance reduces bandwidth requirements for pedestrian detection by over 30x with no deterioration in detection accuracy. Furthermore, this prior work shows that a single static lossy compression tolerance does not work as well on a cloudy or a rainy weather condition as it works in a sunny weather condition. Throughout the day and year environmental conditions change, degrading the utility of a static lossy compression approach. Dynamically adapting the error tolerance could maintain a high pedestrian detection accuracy in adverse environmental situations.

### B. Pedestrian detection and environment classification method

The advent of deep learning significantly improved the accuracy and computational time of object detection and classification. The state-of-the-art deep learning-based object detection models operate in real-time and provide a high detection accuracy. Object detection models are classified into two categories: i) region-based object detection, and ii) single-shot object detection. Region-based object detection models include: Region-Convolutional Neural Network (R-CNN) [17]; Fast R-CNN [18]; and Faster R-CNN [19]. The single-shot object detection models include: Single Shot MultiBox Detector (SSD) [20] and You Only Look Once - Version 3 (YOLOv3) [21]. All these aforementioned deep learning models run in real-time (less than 10 frames per second). However, in terms of pedestrian detection accuracy, YOLOv3 shows a better detection accuracy (81% at 20 fps) [5]. Since safety-critical applications, such as roadway pedestrian detection, require a high detection accuracy, we use YOLOv3 as our pedestrian detection model. In our prior study, we achieve a 98% accuracy for pedestrian detection in sunny weather using YOLOv3 [5].

Deep learning excels in the domain of object and image classification [22]. In the area of deep learning, Convolution Neural Networks (CNNs) excel in image classification tasks [23]. The state-of-the-art CNN-based classification models include: Visual Geometry Group (VGG) [24]; InceptionV3 [25]; and ResNet5 [26]. However, due to the simple architecture and a high classification accuracy on a smaller number of classes, and real-time performance, we use the VGG-16 network to classify the environmental condition. As this convolutional network is 16 layers deep, it is referred to as VGG-16 [24].

## IV. Dynamic Error-Bounded Lossy Video Compression Strategy

Compressing a video with a low-quality level greatly improves the compression ratio and reduces the bandwidth requirement to transfer the video. However, as the quality level of the video decreases, its ability to be used for video analytics decreases as well. For pedestrian detection, this results in lower detection accuracy. Furthermore, environmental conditions (e.g., rain, darkness, fog) alter the compression ratio and makes pedestrian detection more difficult by obscuring pedestrians. Dynamically adapting the video compression quality level based on the current environmental condition ensures that we always detect pedestrians with high accuracy throughout the day and the year.

Figure 1 presents our framework for our dynamic error-bounded lossy compression strategy that uses machine learning to detect pedestrians. This study develops a dynamic feedback control system that adapts the compression level to maintain the same detection accuracy of a system communicating the raw lossless video data. Figure 2 presents a real-world deployment of our dynamic EBLC strategy. In our system, a roadside video monitoring camera collects video data and transfers it to an attached video compression unit. The video compression unit compresses the raw video stream using a set tolerance level. In our experiments, we set the tolerance based on the PSNR ratio between the raw video and the resulting compressed video. The exact PSNR value depends on the environmental conditions (e.g., rain and darkness). We use H.264 for video compression but note that other video compression algorithms work with our dynamic EBLC strategy.

After compression, the compressed video streams are sent wirelessly to the roadside edge computing infrastructure. This edge computing infrastructure contains three main components: i) a set of calibrated pedestrian detection models for different environmental conditions; ii) the active pedestrian detection model to process video image, and iii) an environmental condition detection model to identify the status of the current environment for a given video.

In this research, we use the YOLOv3 machine learning model to detect pedestrians as described in our previous research [5]. (We describe the YOLOv3 model and its calibration process in the subsection B of Section IV.) This paper's study focuses on the development of a dynamic error-bounded video compression strategy that is independent of the vision-based pedestrian detection method. Given an environmental condition, the edge computing infrastructure selects an appropriate model from the set of pre-trained and calibrated models. In addition, it determines the corresponding PSNR for the model that yields the largest reductions in bandwidth while still maintaining the same detection accuracy. The selected PSNR value is periodically sent to the roadside video monitoring camera for use when compressing the video stream. Thus, the video compression unit near the video camera dynamically adapts its compression quality level over time based on the selection of the calibrated machine learning model for the current environmental condition. In our design, a pedestrian detection model is trained with video images with different PSNR values for a specific environmental condition. For example, we select three types of rain: 1) light rain; 2) moderate rain; and 3) heavy rain. For each level of rain, we initially compress the video to six different PSNR levels measured in decibels (dB): (10) 56 dB; (20) 49 dB; (30) 43 dB; (40) 37 dB; (50) 31 dB; and (51) 30 dB. The value in the parentheses shows the Constant Rate Factor (CRF) corresponding to each PSNR value. The CRF is the error control knob we tune for the compressors inside FFmpeg [27], a software tool used to process audio and video files.

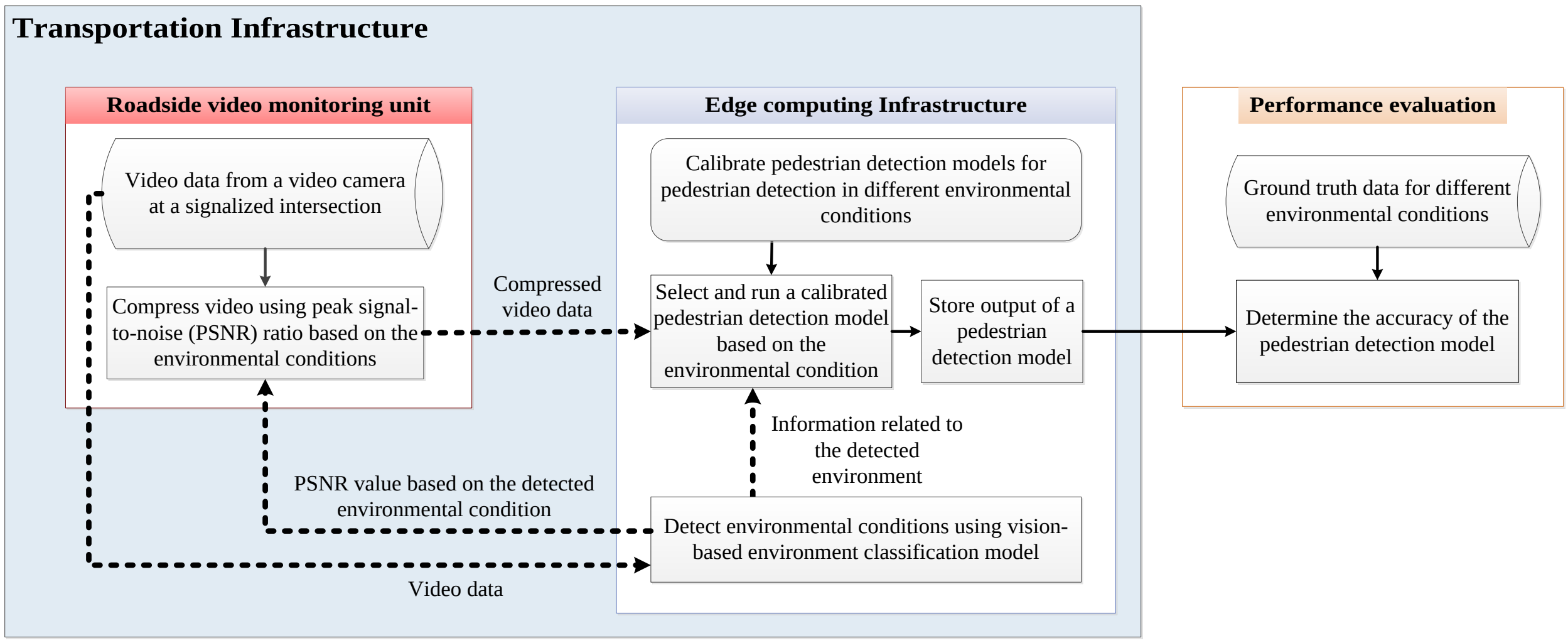

Fig. 1. Dynamic error-bounded lossy video compression strategy.

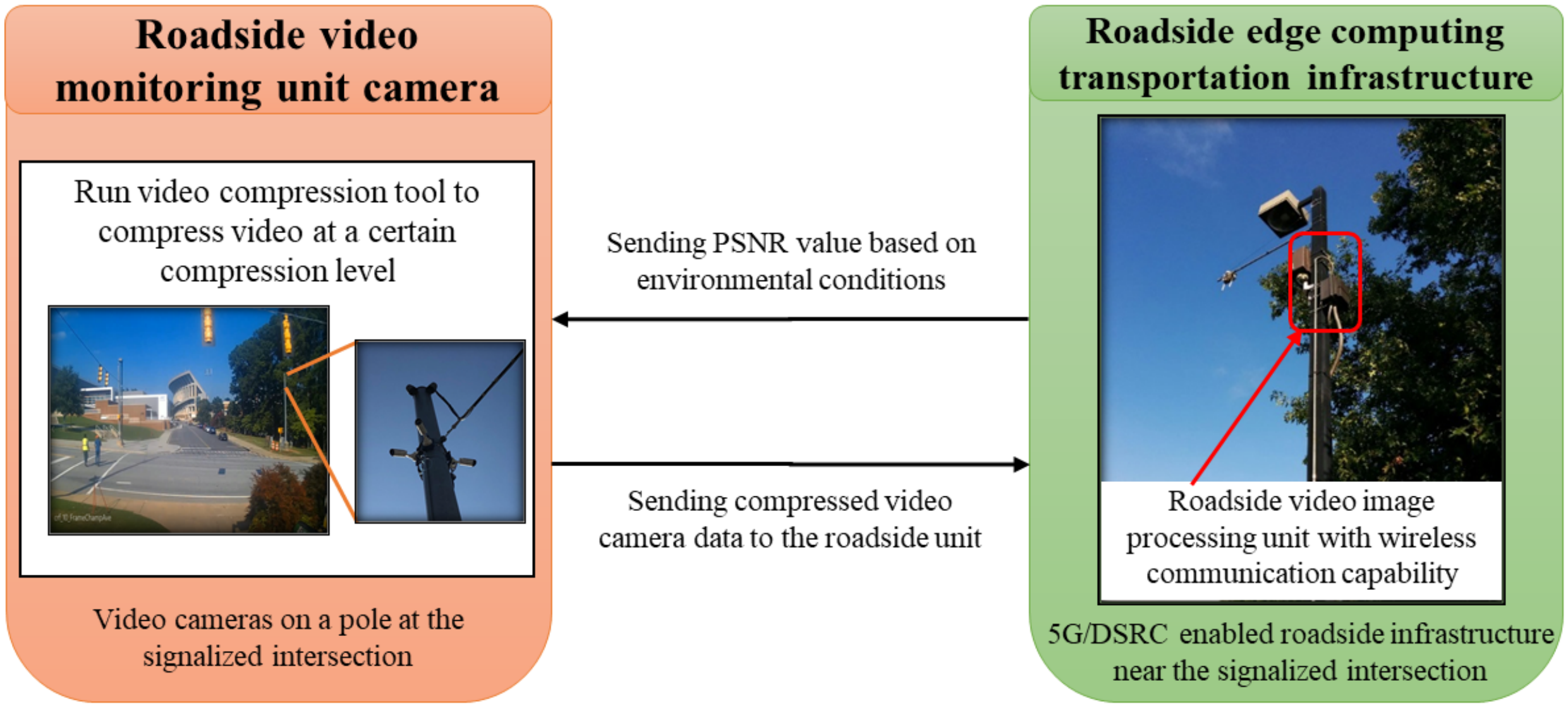

Fig. 2. Real-world implementation scenario of the dynamic EBLC strategy.

To determine the optimum CRF and corresponding PSNR that maintains a high pedestrian detection accuracy, a reference table is constructed offline. The reference table contains only the models that have a pedestrian detection accuracy equal to that of the baseline model. To construct the reference table and the catalog of corresponding models, we train and evaluate a model on data compressed with a CRF of 10 (highly accurate) along with computing the PSNR. Next, we increase the CRF by 10 (degrading video quality and improving compression) until the new model's detection accuracy drops below the minimum threshold. At this point, we vary the CRF by 1 to fully explore the range between the last valid CRF and the first invalid CRF. Again, we evaluate each model to determine if it meets our quality-of-service standards; rejecting any models that do not. After exploring each CRF in the interval, we have a table that allows us to select a trained model given a requested CRF or PSNR value.

We calculate the accuracy of the pedestrian detection model by comparing it with manually annotated ground truth data. To establish a baseline accuracy, we perform pedestrian detection on the uncompressed video feed coming from traffic cameras for all scenarios and calculate the accuracy based on a manually annotated ground truth. For a compression baseline, we compress the video stream to a fixed quality level using standard image difference metric, PSNR, and use a pedestrian detection model with weights calibrated for the compressed data.

In this compression framework, there are three steps: i) lossy video compression; ii) calibration of the pedestrian detection model; and iii) environmental condition detection using an environment classification model. The following subsections describe, in detail, our approach for each step in our dynamic error-bounded lossy compression technique.

### A. *Lossy Video Compression*

Using field-collected data, we compress each video using different Constant Rate Factor (CRF) values using the FFmpeg video compression tool [27]. The video compression level is controlled by the CRF value, and the CRF range is from 0 to 51; where 0 indicates no compression, and 51 indicates maximum compression level. After that, we calculate the PSNR by comparing the original video file and compressed video file.

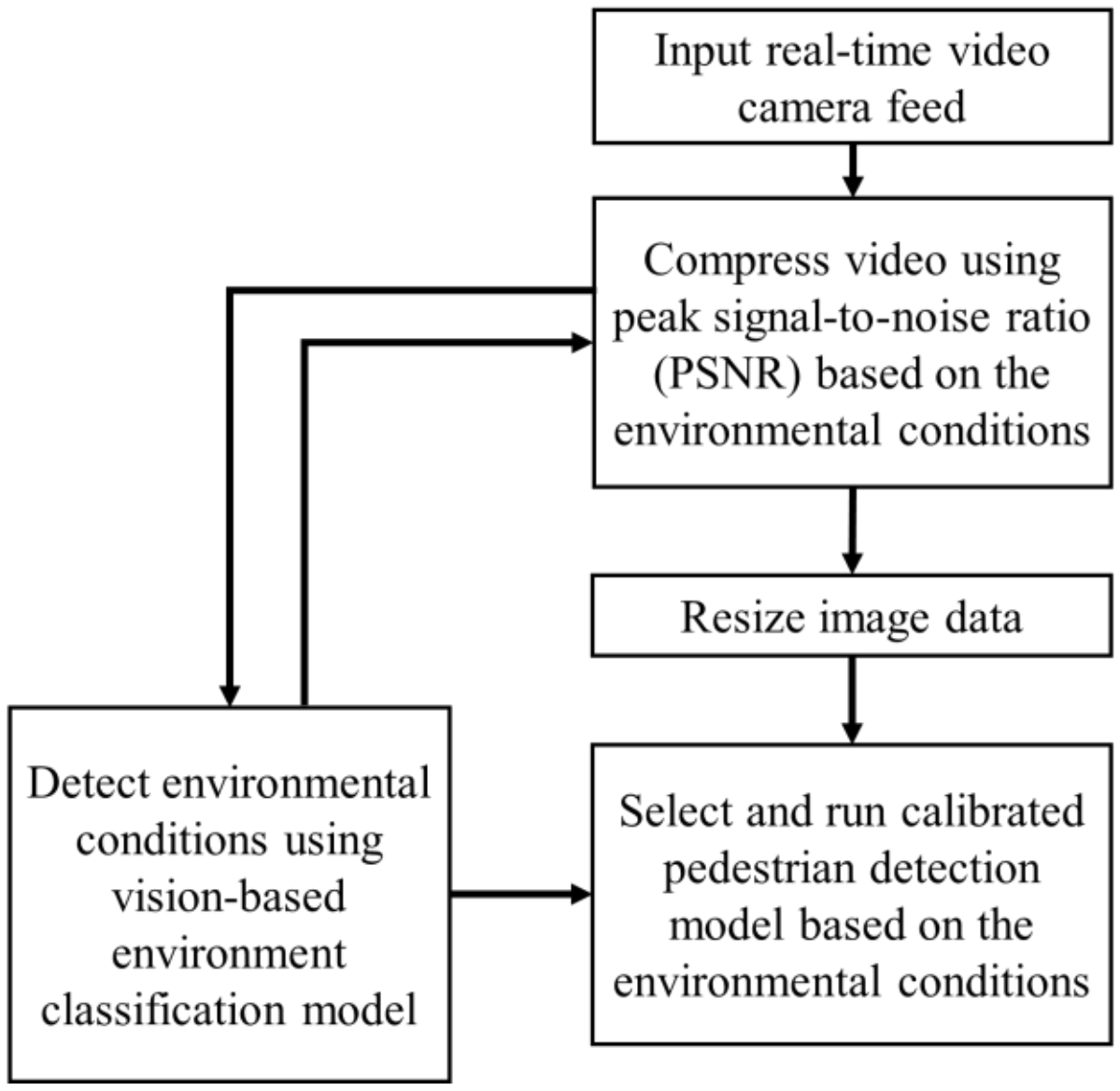

Fig. 3. Feedback-based real-time EBLC algorithm.

Thus, we use the CRF of FFmpeg to compress videos yielding different compression ratios (i.e., small for CRFs near 0 and large for CRFs near 51). However, to make our results independent from the FFmpeg tool, we determine the PSNR value corresponding to each CRF value. Figure 3 presents the feedback-based EBLC algorithm, which compresses the video feed based on the environmental condition. The compression tool compresses the video to a compression level, which maintains a high pedestrian detection accuracy. After resizing the image of the compressed video, a detection model is selected from a library of calibrated pedestrian detection models that account for various environmental conditions.

### *B. YOLOv3 Model Calibration*

The YOLOv3 model [21] divides an image into multiple regions and assigns probabilities to the bounding boxes for each region where a feature is detected. This model can capture the global context of the image as it looks at the whole image simultaneously. The YOLOv3 model consists of 24 convolutional layers followed by 2 fully connected layers and 1×1 reduction layer followed by 3×3 convolutional layers [21]. The YOLOv3 model can have different input image sizes, such as 320×320×3, 416×416×3 and 608×608×3. Based on our experiments, we found that the input image size of 416×416×3 provides the highest pedestrian detection accuracy with a low computational cost. In this study, we use the input image size of 416×416×3 and then normalized the image at the preprocessing layer of the YOLOv3 model.

To achieve a much higher pedestrian detection accuracy for different environmental conditions (e.g., rain and lighting), we train the YOLOv3 model on augmented data. We perform data augmentation for different rain and lighting conditions to produce more realistic images for darkness and rain. To generate augmented data for the YOLOv3 model calibration, we alter the darkness of the images by changing the pixel values of the first channel in the HSL (hue, saturation, lightness) color space of an image. Based on the rain intensity, different types of rainy environments are created by adding random small lines on the image and making the image a little blurry to replicate a realistic rainy environment [28].

To train the YOLOv3 model, we down sample the video at 10 frames per second (fps) to extract frames for pedestrian safety applications [29]. After that, we have used the standard Pascal Visual Object Class (VOC) format to annotate each extracted frame from the video file. The YOLOv3 model splits an image into many regions and calculates the probabilities for each region. Based on the calculated probabilities, the model generates bounding boxes for pedestrians. The YOLOv3 model may generate multiple bounding boxes for a pedestrian, which will reduce the pedestrian detection accuracy significantly. We have used a non-max suppression method [30] to improve the pedestrian detection accuracy by keeping one bounding box and excluding other unnecessary bounding boxes detecting each pedestrian.

### *C. Environmental Condition Detection*

In this study, we use a vision-based Convolution Neural Network (CNN) deep learning model to detect and classify different environmental conditions. The classifier takes an image as input and classifies it among seven different environmental conditions: normal weather, light dark, medium dark, high dark, light rain, moderate rain, and heavy rain (as shown in Table I). The input image is of size 416×416×3 pixels, and the output is a 7×1 matrix, W. For example, as shown in Figure 4, an output, W = [0, 0, 1, 0, 0, 0, 0] indicates a medium dark weather condition. Being a simple CNN-based classifier, the model is able to run on a roadside edge computing device.

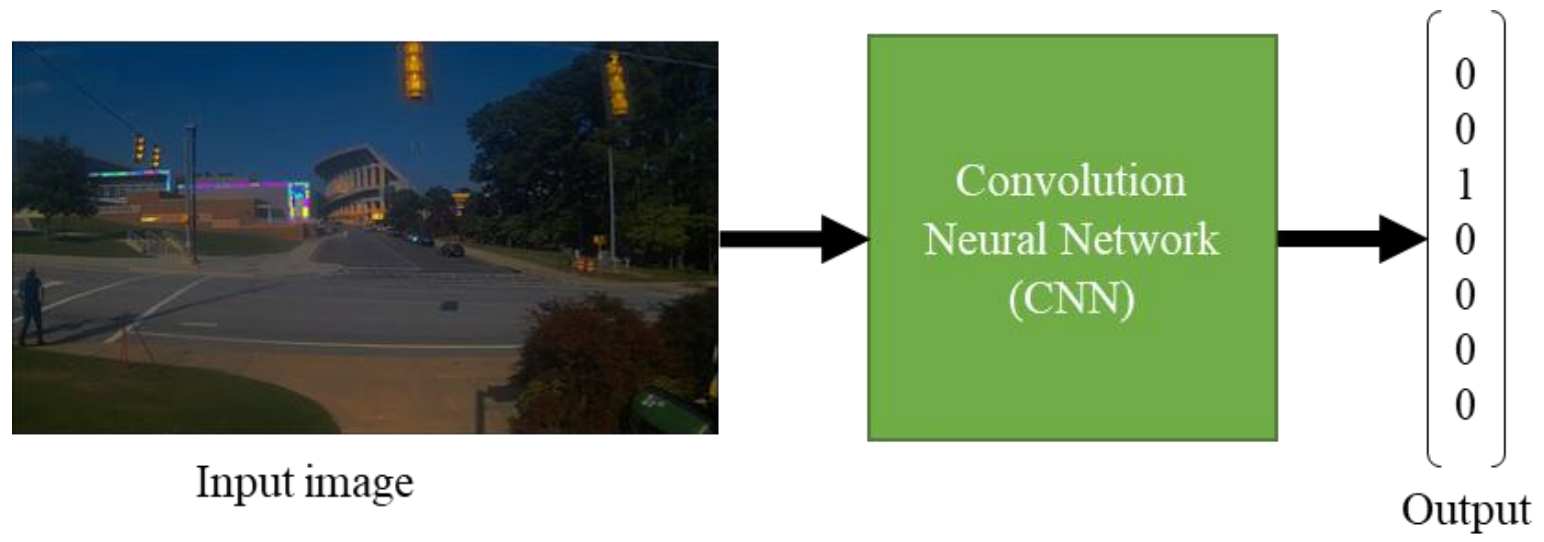

Fig. 4. Environmental condition classifier using Convolution Neural Network (CNN).

TABLE I
SELECTED ENVIRONMENTAL CONDITIONS AND CORRESPONDING VIDEO COMPRESSION SCENARIOS

| **Evaluation scenarios** | **Environmental condition** | **Category of environmental condition** | **Constant Rate Factor (CRF) Range** | **Minimum Average PSNR value corresponding to CRF value in Column 4** |
|---|---|---|---|---|
| 1 | Normal | Sunny weather | 0-10, 11-20, 21-30, and 31-33 | 56 dB (corresponding to CRF 10), 49 dB (corresponding to CRF 20), 43 dB (corresponding to CRF 30), and 41dB (corresponding to CRF 33) |
| 2 | Lighting condition | Light dark | | |
| 3 | | Medium dark | | |
| 4 | | High dark | | |
| 5 | Rainy condition | Light rain | | |
| 6 | | Moderate rain | | |
| 7 | | Heavy rain | | |

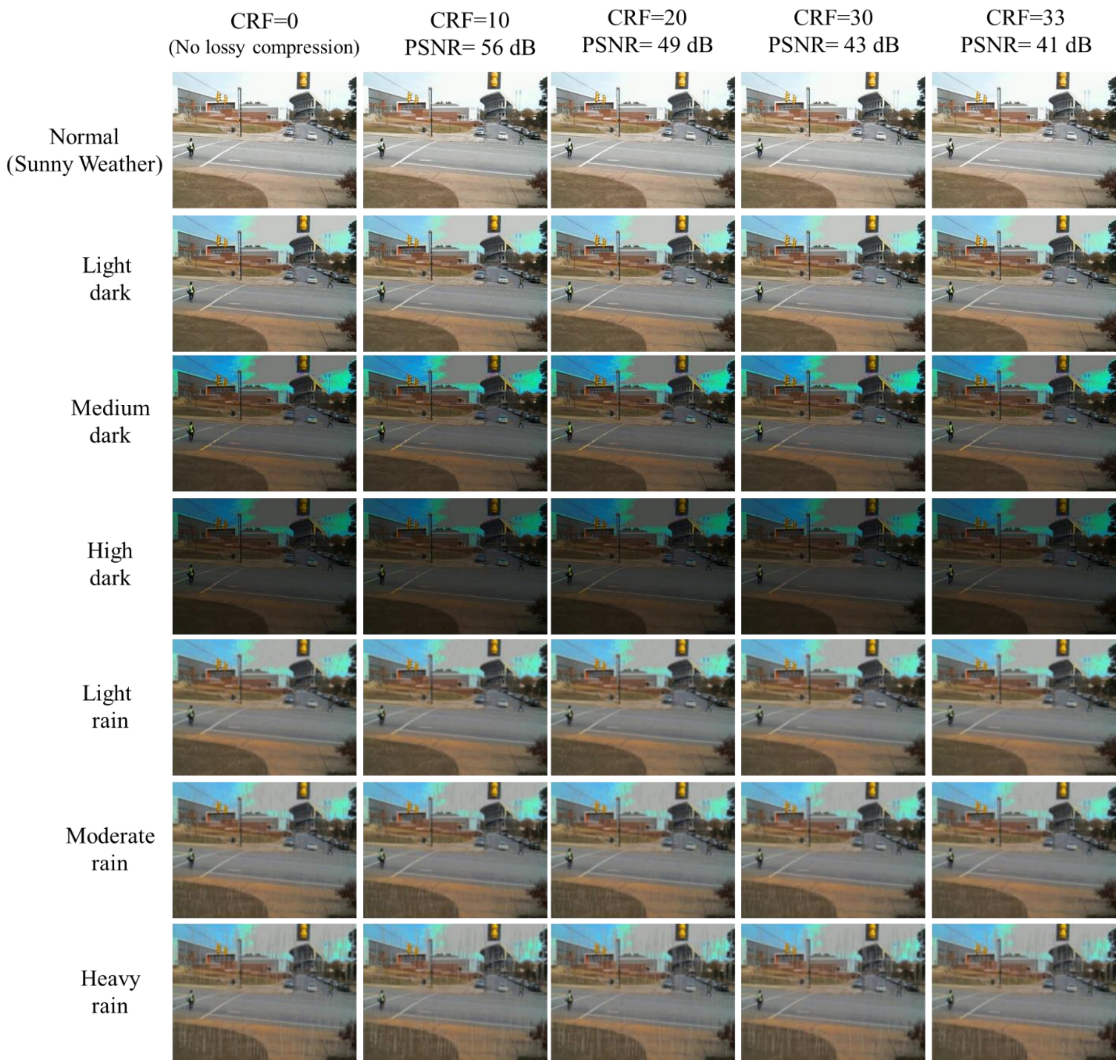

Fig. 5. Video compression with different environmental weather conditions.

## V. Analysis and Results

In this section, we describe the environmental and lossy video compression scenarios, data generation and deep learning model calibration for different environmental conditions. In addition, we report the pedestrian detection accuracy for each condition.

### A. *Environmental and Lossy Video Compression Scenarios*

In this study, we consider three different environmental conditions: (i) normal condition (i.e., sunny weather); (ii) lighting condition; and (iii) rainy condition. For the lighting and rainy conditions, we further break these down into three additional categories. The categories for the lighting condition are light, medium and high, and the categories for the rainy condition are light, moderate and heavy. Prior work found that the pedestrian detection accuracy for the sunny weather condition decreases from the baseline condition (no compression) if the CRF value is greater than 30 (where PSNR value is 43dB) [5]. Thus, for each category of environmental condition, for space reasons in the paper, we presented four compression scenarios: (a) CRF = 10 (PSNR=56 dB); (b) CRF = 20 (PSNR=49 dB); (c) CRF = 30 (PSNR=43 dB); (d) CRF = 33 (PSNR=41 dB). However, in a real-world deployment, more models would be used. After collecting video data for the normal weather condition, we generate data for the different environmental conditions and compression scenarios to evaluate pedestrian detection accuracy. For each scenario, we calculate the pedestrian detection accuracy to determine the maximum compression ratio at which we maintain the baseline pedestrian detection accuracy.

### B. Data Generation and Description

To obtain data for our baseline normal sunny weather condition (no data compression), we collect field data from the Perimeter Road and Avenue of Champions intersection at Clemson, South Carolina. We use a camera on a data collection pole and record video of the intersection including pedestrians on the crosswalk. This dataset contains a total of 427 images where pedestrians are moving in four directions, such as north-south, south-north, east-west and west-east. After collection of the field data, we perform data augmentation to generate images for different environmental conditions. Using data augmentation as described in the pedestrian detection model calibration subsection of Section IV, we create seven environmental conditions as shown in Figure 5. Thus, for each environmental condition, we generate 427 images based on the data collected from the field. This dataset is publicly available at https://drive.google.com/open?id=1XA0hOfjvIb1l29rvkbUnwjN6kMj12KaD.

### C. YOLOv3 Model Calibration

To improve the pedestrian detection accuracy using the YOLOv3 model, we use a pre-trained model and retrained the model on our collected and generated dataset for different environmental conditions and different compression levels. For the normal sunny weather's 427 images, we split our dataset further into train, test, and validation sets with the following percentages 63%, 20%, and 17%, respectively. In total, we evaluate 28 unique configurations of the compression level. After data augmentation, we generate a total of 14,945 images, which includes 9,415 images for training, 2,989 images for testing, and 2,540 images for validation. Using these datasets, we retrain the YOLOv3 model. After training, to further improve the pedestrian detection accuracy, we use a non-max suppression method to suppress false positives in the pedestrian detection output.

### D. Environmental Condition Detection

To detect different environmental conditions, we use a CNN-based deep learning model. In particular, we use the VGG-16 model [24] as the base network, and we train on our own dataset. We normalize and resize each input image from 416×416× 3 to a size of 224 × 224 × 3 to match with the VGG-16 model input layer size. The convolution network with linear rectified units (ReLU), max pooling, and a fully connected layer with ReLU acts as an image encoder to extract the image features of various weather conditions. As shown in Figure 6, this classification model classifies the image into one of the seven classes: normal-sunny weather, light dark, medium dark, high dark, light rain, moderate rain, and heavy rain. The model is trained on the augmented datasets for these seven environmental conditions. Similarly, we split the datasets into 63% for training, 20% for testing, and 17% for validation for each weather condition. Based on the testing dataset, our CNN-based model is able to classify the weather condition with 97% accuracy.

### E. Evaluation of Dynamic EBLC Framework

To investigate the impact of different environmental conditions on pedestrian detection accuracy, we evaluate the accuracy of the YOLOv3 model for pedestrian detection in different weather conditions by training only on sunny weather data. Figure 7 presents pedestrian detection accuracy for different environmental conditions with no data compression. We found that the pedestrian detection accuracy continues to reduce as the weather condition continues to deteriorate. Pedestrian detection accuracy will reduce even more if we compress the image before pedestrian detection during adverse weather conditions. Thus, the accuracy of the pedestrian detection model varies based on the environmental condition and the degree lossy compression, and it is important to train the pedestrian detection model with data for different environmental conditions and on the level of lossy compression.

We evaluate the pedestrian detection accuracy for different environmental conditions with different CRF values ranging from 0 to 33. We limited our CRF value to 33 as, CRF values above 33 (41dB) yields unacceptable deterioration in the pedestrian detection accuracy. Figure 8 shows that using our dynamic EBLC framework, the pedestrian detection accuracy for the different weather conditions has improved, and in heavy rain condition, we found a 14% improvement compared to the baseline condition, i.e., no lossy compression, as shown in Table II. As the weather condition becomes more adverse, the pedestrian detection accuracy goes down with the degradation of weather conditions. Similarly, the detection accuracy decreases as the CRF values increase, meaning that the pedestrian detection accuracy decreases as the compression ratio increases. As presented in Figure 8, we can determine the minimum video quality level that still maintains a fixed pedestrian detection threshold. In our case, we consider the pedestrian detection baseline accuracy as 97%, which is considered as the lowest accuracy we found for all environmental scenarios without any data compression (CRF=0).

Table II shows the maximum CRF or minimum PSNR based on the different environmental conditions. The original communication bandwidth without any compression is 9.82 MBits/sec. The maximum CRF ranges from 0 to 30 while the minimum PSNR ranges from 41dB to 56dB. Depending on the PSNR value, we reduce bandwidth by 1.5× to 18× in our case study. However, our previous study [5] showed that using EBLC and a static error tolerance reduces bandwidth requirements for pedestrian detection by over 30x with no deterioration in detection accuracy. As stated in the literature review section of the paper, the static tolerance does not work well on adverse weather conditions, such as in cloudy or rainy conditions as was the case in our current study. Because we use a different data set and we compress data to a fixed-accuracy (i.e., target PSNR) in the current study, it allows for variations in the compression ratio in order to meet a fixed-accuracy requirement.

TABLE II
MAXIMUM COMPRESSION RATIO ACHIEVED FOR DIFFERENT WEATHER CONDITIONS

| Weather condition | Baseline Pedestrian Detection Accuracy (no compression) | Dynamic EBLC Framework Pedestrian Detection Accuracy | Improvement in Pedestrian Detection using Dynamic EBLC Framework | Maximum Constant Rate Factor (CRF) (or Minimum PSNR) | Required Bandwidth (MBits/sec) | Bandwidth Reduction |
|---|---|---|---|---|---|---|
| **Normal** | 98% | 97% | -1% | 30 (41dB) | 0.53 | 18× |
| **Light Dark** | 93% | 97% | 4% | 30 (41dB) | 0.68 | 14× |
| **Medium Dark** | 92% | 97% | 5% | 20 (43dB) | 1.01 | 9.5× |
| **High Dark** | 90% | 97% | 7% | 10 (56dB) | 4.05 | 2.5× |
| **Light Rain** | 89% | 97% | 8% | 10 (56dB) | 5.15 | 1.5× |
| **Medium Rain** | 88% | 97% | 9% | 0 | 9.82 | 0× |
| **Heavy Rain** | 83% | 97% | 14% | 0 | 9.82 | 0× |

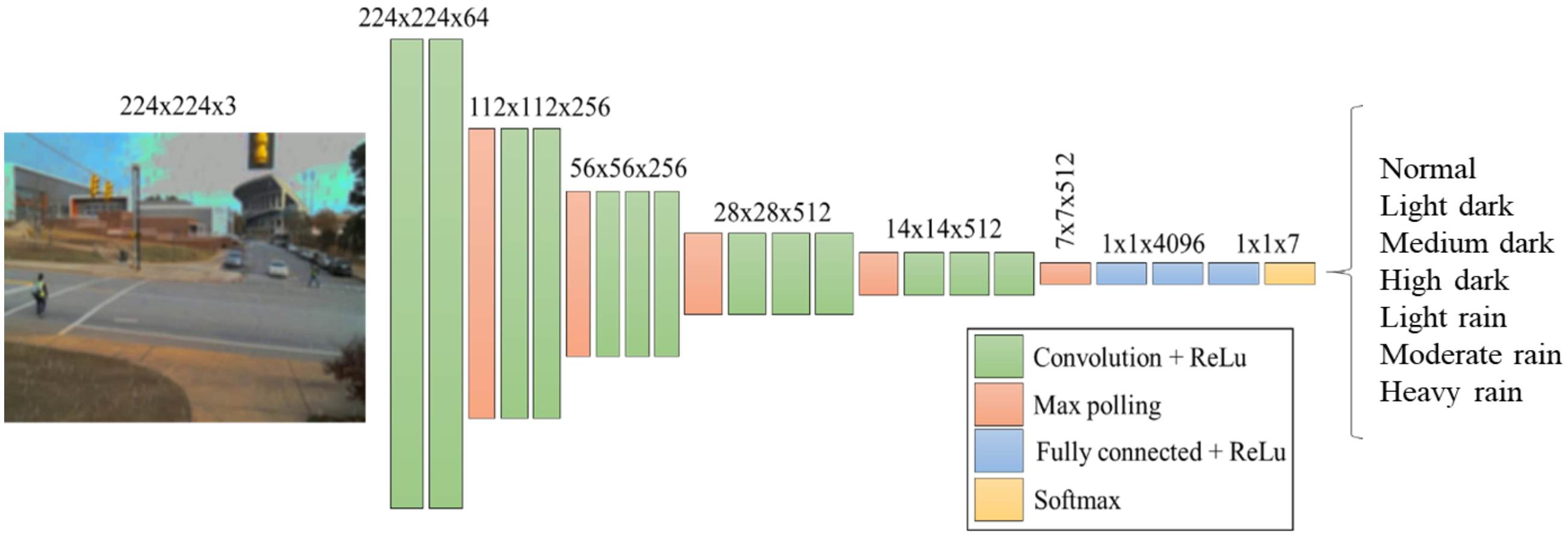

Fig. 6. CNN-based environmental condition classifier.

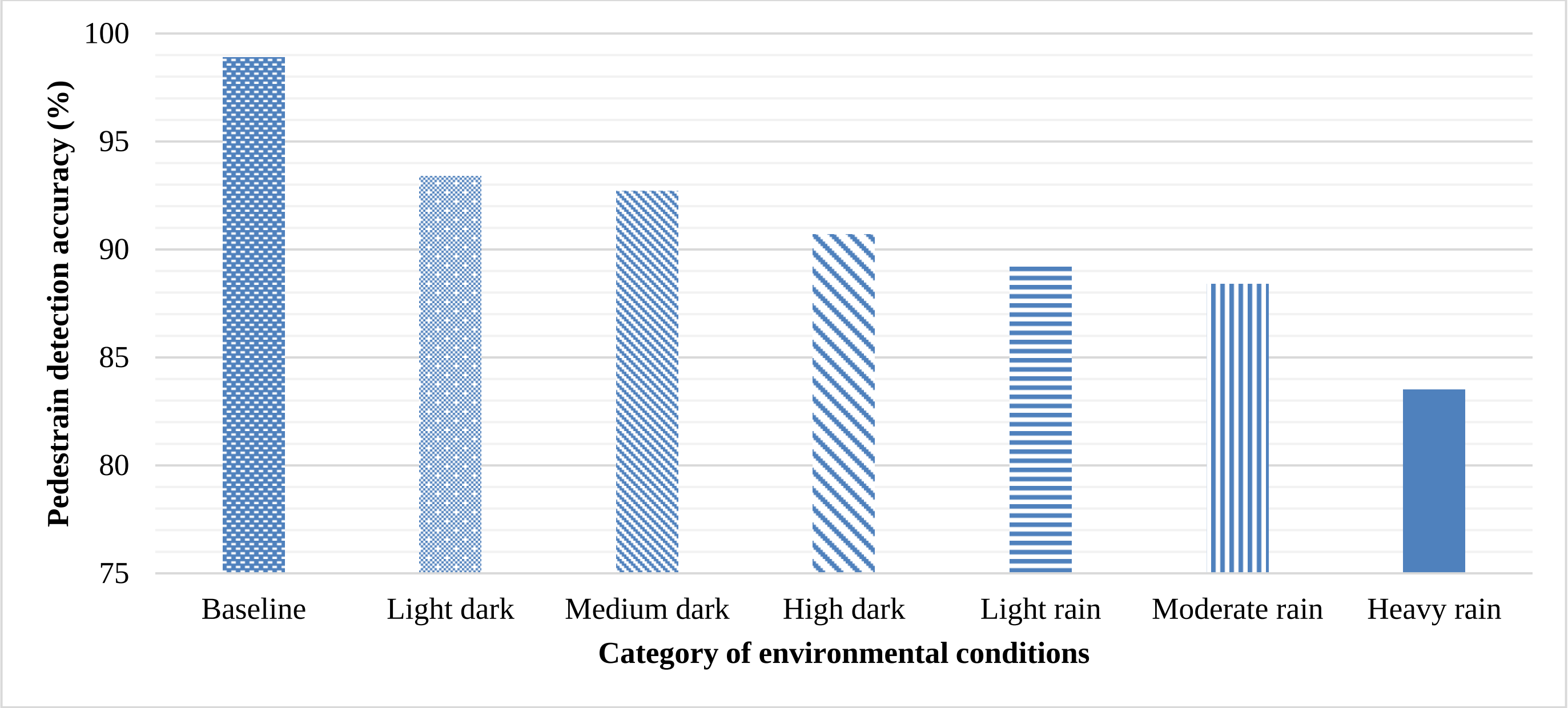

Fig. 7. Pedestrian detection accuracy for different environmental conditions with no data compression.

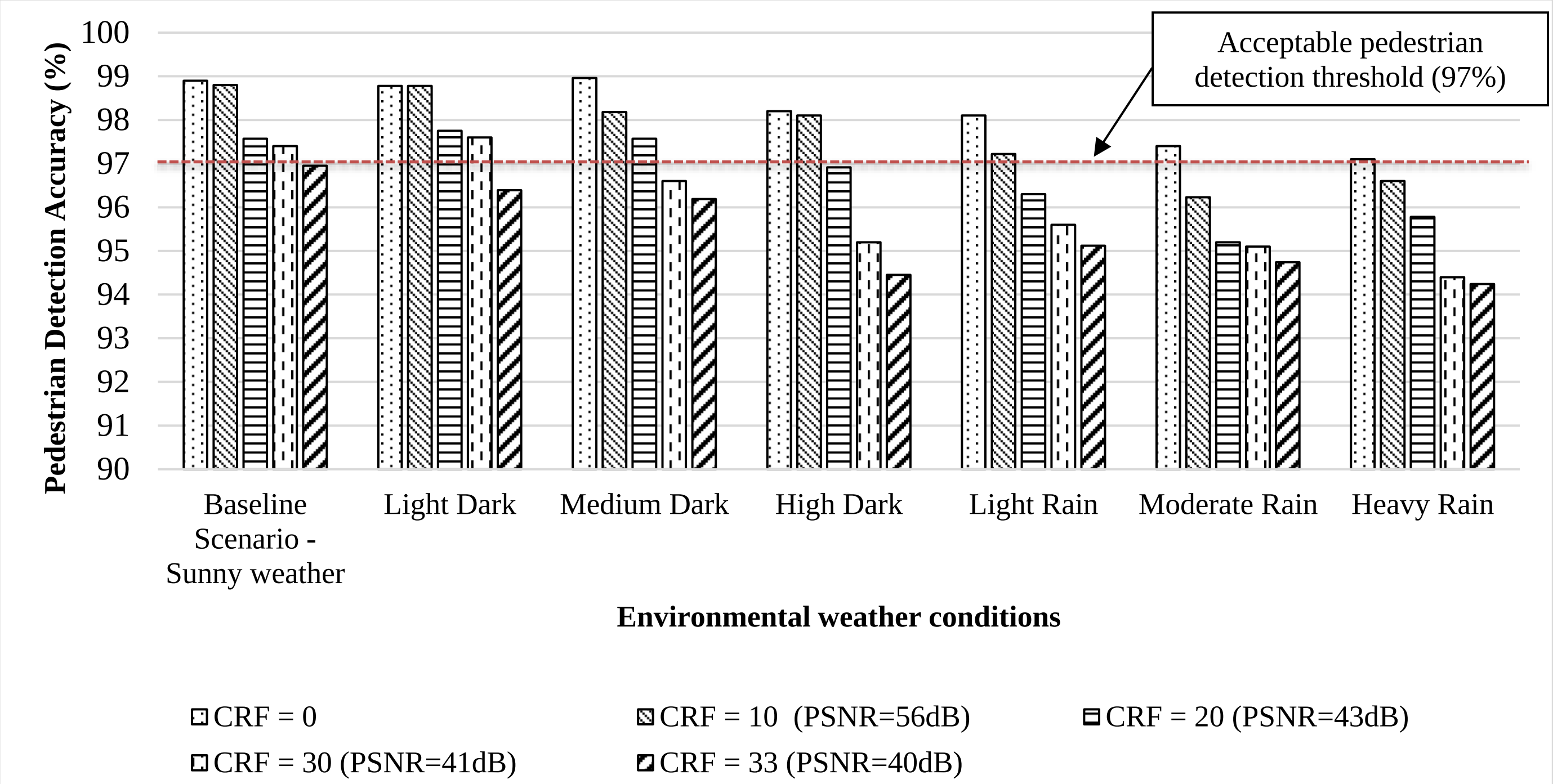

Fig. 8. Pedestrian detection accuracy for different weather conditions with different compression levels

## VI. Conclusion

Dynamically adapting the video compression quality level based on environmental conditions ensures the reduction of the communication bandwidth requirement for transferring a video wirelessly while detecting pedestrians with high accuracy. The contribution of this study is to develop a feedback-based real-time dynamic EBLC strategy considering different environmental conditions by reducing the communication bandwidth while maintaining a baseline (no compression and sunny weather) pedestrian detection accuracy. Depending on different environmental factors, our strategy dynamically selects the error tolerance for error-bounded lossy compression that yields the best performance. Through our dynamic EBLC strategy, we maintain a high pedestrian detection accuracy using the YOLOv3 detection model across a selection of the different environmental levels of rain and darkness. Our EBLC strategy is independent of the pedestrian detection model, and any type of pedestrian detection model can be used in our framework. Our analysis reveals that in adverse environmental conditions, the dynamic EBCL strategy can reduce the bandwidth requirements for transmitting video over prior approaches up to 14x while maintaining the baseline accuracy that transmits lossless videos. Results show that if the weather condition is adverse, the bandwidth reduction is lower. Even for moderate and heavy rainy conditions, we could not compress video at all if we are required to maintain a 97% pedestrian detection accuracy. In our future study, we will consider unexplored trade-offs, such as the energy efficiency of our ELBC strategy and how to utilize multiple intra-frame compression tolerances to further improve the compression ratio to maximize the bandwidth usage.

## Acknowledgments

This material is based on a study partially supported by the Center for Connected Multimodal Mobility ($C^2M^2$) (USDOT Tier 1 University Transportation Center) Grant headquartered at Clemson University, Clemson, South Carolina, USA. Any opinions, findings, and conclusions or recommendations expressed in this material are those of the author(s) and do not necessarily reflect the views of the Center for Connected Multimodal Mobility ($C^2M^2$), and the U.S. Government assumes no liability for the contents or use thereof. This material is also based upon work supported by the National Science Foundation under Grant No. SHF-1910197.

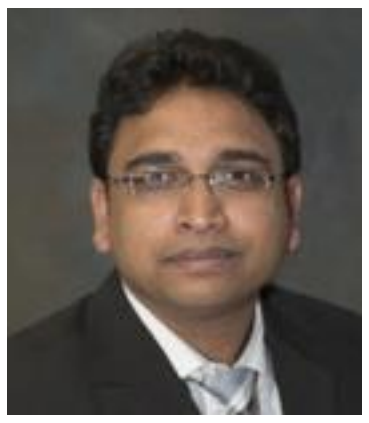

**Mizanur Rahman** received the M.Sc. and Ph.D. degrees, majoring in civil engineering with transportation systems, from Clemson University, in 2013 and 2018, respectively. Since 2019, he has been an Assistant Director with the Center for Connected Multimodal Mobility ($C^2M^2$), a U.S. Department of Transportation Tier 1 University Transportation Center (cecas.clenson.edu/c2m2), Clemson University. He was closely involved in the development of Clemson University Connected and Autonomous Vehicle Testbed. His research focuses on transportation cyber-physical systems for connected and autonomous vehicles and for smart cities.

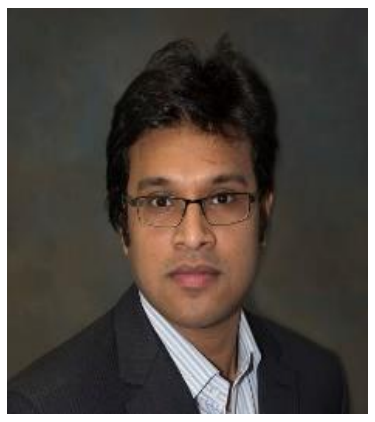

**Mhafuzul Islam** received the BS degree in Computer Science and Engineering from the Bangladesh University of Engineering and Technology in 2014 and MS degree in Civil Engineering from Clemson University at 2018. He is currently a Ph.D. student in the Glenn Department of Civil Engineering at Clemson University. His research interests include Transportation Cyber-Physical Systems with an emphasis on Data-driven Connected Autonomous Vehicle. He is a student member of IEEE.

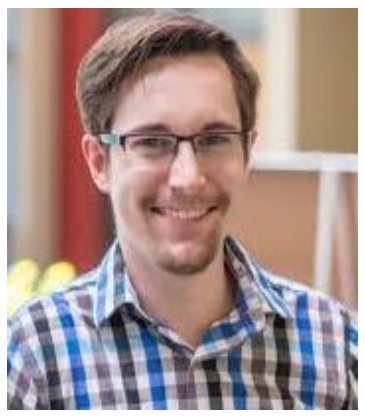

**Jon C. Calhoun**, Ph.D. Jon Calhoun is an Assistant Professor in the Holcombe Department of Electrical and Computer Engineering at Clemson University. He received a B.S. in Computer Science from Arkansas State University in 2012, a B.S. in Mathematics from Arkansas State University in 2012, and a Ph.D. in Computer Science from the University of Illinois at Urbana-Champaign in 2017. His research interests lie in fault tolerance and resilience for high-performance computing (HPC) systems and applications, lossy and lossless data compression algorithms and their use in HPC and intelligent transportation systems, scalable numerical algorithms, power-aware computing, and approximate computing.

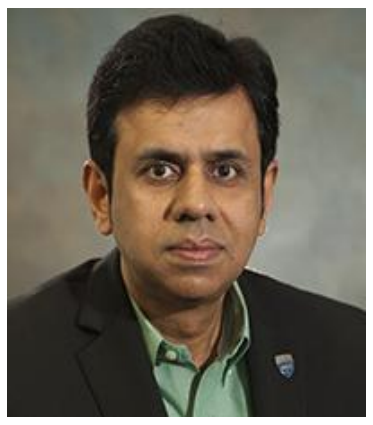

**Mashrur Chowdhury** (SM'12) received the Ph.D. degree in civil engineering from the University of Virginia, USA in 1995. Prior to entering academia in August 2000, he was a Senior ITS Systems Engineer with Iteris Inc. and a Senior Engineer with Bellomo–McGee Inc., where he served as a Consultant to many state and local agencies, and the U.S. Department of Transportation on ITS related projects. He is the Eugene Douglas Mays Professor of Transportation with the Glenn Department of Civil Engineering, Clemson University, SC, USA. He is also a Professor of Automotive Engineering and a Professor of Computer Science at Clemson University. He is the Director of the USDOT Center for Connected Multimodal Mobility (a TIER 1 USDOT University Transportation Center). He is Co-Director of the Complex Systems, Data Analytics and Visualization Institute (CSAVI) at Clemson University. Dr. Chowdhury is the Roadway-Traffic Group lead in the Connected Vehicle Technology Consortium at Clemson University. He is also the Director of the Transportation Cyber-Physical Systems Laboratory at Clemson University. Dr. Chowdhury is a Registered Professional Engineer in Ohio, USA. He serves as an Associate Editor for the IEEE TRANSACTIONS ON INTELLIGENT TRANSPORTATION SYSTEMS and Journal of Intelligent Transportation Systems. He is a Fellow of the American Society of Civil Engineers and a Senior Member of IEEE.